\documentclass{sig-alternate}
\usepackage{amssymb}
\usepackage{amsmath}
\usepackage{pgfplots}
\usepackage[ruled,vlined]{algorithm2e}
\usepackage{algpseudocode}

\newtheorem{definition}{Definition}

\begin{document}

\setcopyright{acmcopyright}
\conferenceinfo{GECCO '15,}{July 11 - 15, 2015, Madrid, Spain}
\isbn{978-1-4503-3472-3/15/07}\acmPrice{\$15.00}

\doi{http://dx.doi.org/10.1145/2739480.2754685}
%
%
%
    \clubpenalty=10000
    \widowpenalty = 10000

\title{Theoretical Perspective of Convergence Complexity of Evolutionary Algorithms Adopting Optimal Mixing}

\numberofauthors{2}
\author{
\alignauthor
Yu-Fan Tung\\
       \affaddr{Taiwan Evolutionary Intelligence Laboratory}\\
       \affaddr{Department of Electrical Engineering}\\
       \affaddr{National Taiwan University}\\
       \email{r02921044@ntu.edu.tw}
\alignauthor
Tian-Li Yu\\
       \affaddr{Taiwan Evolutionary Intelligence Laboratory}\\
       \affaddr{Department of Electrical Engineering}\\
       \affaddr{National Taiwan University}\\
       \email{tianliyu@ntu.edu.tw}
}
\date{\today}

\maketitle
\begin{abstract}

The optimal mixing evolutionary algorithms (OMEAs) have recently drawn much attention for their robustness, small size of required population, and efficiency in terms of number of function evaluations (NFE).
In this paper, the performances and behaviors of convergence in OMEAs are studied by investigating the mechanism of optimal mixing (OM), the variation operator in OMEAs, under two scenarios---one-layer and two-layer masks.
For the case of one-layer masks, the required population size is derived from the viewpoint of initial supply, while the convergence time is derived by analyzing the progress of sub-solution growth.
NFE is then asymptotically bounded with rational probability by estimating the probability of performing evaluations.
For the case of two-layer masks, empirical results indicate that the required population size is proportional to both the degree of cross competition and the results from the one-layer-mask case.
The derived models also indicate that population sizing is decided by initial supply when disjoint masks are adopted,
that the high selection pressure imposed by OM makes the composition of sub-problems impact little on NFE,
and that the population size requirement for two-layer masks increases with the reverse-growth probability.

\end{abstract}

\category{F.2.m}{Analysis of Algorithms and Problem Complexity}{Miscellaneous}
\category{I.2.8}{Artificial Intelligence}{Problem Solving, Control Methods, and Search}[Heuristic methods]


\terms{Experimentation, Performance, Theory}

\keywords{
Optimal Mixing;
Convergence Complexity;
Population Sizing;
Convergence Time;
Number of Function Evaluations}

\section{Introduction}
\label{sec:introduction}

The importance of convergence analysis of evolutionary algorithms cannot be overemphasized.
Without theoretical support, the empirical findings cannot be generalized,
and the development of new mechanism lacks direction.
However, due to the stochastic nature, the analyses of evolutionary algorithms (EAs)  are difficult,
and hence are either inaccurate and incomplete or come much later than the debut of the algorithm.
For example, the concept of the simple GA was proposed around the late 1960s.
Accurate facet-wise models for the population sizing and the convergence time were not proposed until the late 1990s~\cite{1992_Goldberg, 1993_Thierens, 1993_BGA, 1999_Gamblers, 2001_Supply}.
Similarly, the technique of model building in estimation of distribution algorithms (EDAs) has been developed since the late 1990s,
and complexity analysis of model building, the kernel of EDA, appeared in the field about ten years later~\cite{2002_Pelikan, 2007_Yu}.

Recently, the optimal mixing operator (OM), proposed by Thierens and Bosman,
has drawn much attention for its robustness and performance~\cite{2012_LN_OM_FI, 2013_Hier}.
Utilizing the intermediate results during variation, mixing with OM is considered noise-free decision making, which greatly reduces the required population size and boosts the performance.
The performance of OM significantly depends on the linkage sets~\cite{2012_LN_OM_FI,2012_Goldman}.
Recent researches involve using the CP index~\cite{2014_CP} and filtering/combining linkage hierarchies~\cite{2013_Filtering} to decide whether a given mask is promising during optimization.
However, those techniques have been developed with little theoretic support.

The goal of this paper is to analyze the convergence complexity of EAs adopting OM,
also called OMEAs, when various linkage models are adopted.
In particular, we focus on deriving analytical models for
the required population size, the convergence time, and the number of function evaluations (NFE).
%
%
Among many OMEAs, the analyses in this paper suit best the scenario of GOMEA~\cite{2011_OMEA}
since our study heavily uses probabilistic models.

For the rest of this paper, the background is first briefly introduced in Section~\ref{sec:relatedWorks}.
In the scenario of one-layer masks, complexity models are proposed in Section~\ref{sec:nonOverlappingMasks}.
The results are then extended to two-layer masks in Section~\ref{sec:twoLayerNonOverlappingMasks},
followed by conclusion.

\section{Background}
\label{sec:relatedWorks}

In this section, background knowledge of this paper is provided.
Since the behavior of OMEAs depends on the linkage sets,
we first formally define the family of subsets and the associated notations.
Using the notations, the test problems used in this paper are then described.
We then give a brief introduction to GOMEA, which is the research scenario of this paper.
Finally, since the concept of initial supply is highly related to our work, it is also addressed.

%


\subsection{Family of Subsets}

In OMEAs, variables are mixed according to linkage sets
to ensure adequate mixing while preventing loss of promising partial solutions.
The family of subsets~\cite{2010_LTGA}, denoted by $\mathcal F$, is a general model of linkage sets.
Before defining the FOS we introduce the necessary notations.
We denote the length of chromosome by $\ell$, and the population size by $n$.
Variables in one chromosome, also called genes, can be expressed in a vector $\vec{x} = \langle x_1, \dots, x_\ell\rangle$.
The notation $|{\vec x}|$ denotes the number of elements in $\vec x$.
The natural number set is denoted by $\mathbb N$.
Using these notations the FOS is defined as follows.

\begin{definition}
For any integer $\ell\in{\mathbb N}$, $S_\ell$ is an {\bf index set} which consists of all integers from $1$ to $\ell$.
Subsets of $S_\ell$ are called {\bf masks}.
\begin{equation*}
S_\ell = \{ 1, 2, \dots, \ell\}.
\end{equation*}
\end{definition}

\begin{definition}
For an index set $S_\ell$, $\mathcal F$ is a {\bf{family of subsets (FOS)}} with the following two properties.
\begin{enumerate}

\item $\mathcal F$ is an ordered set consisting of masks.
\begin{equation*}
\begin{split}
{\mathcal F} = \langle {\mathbf F}^1, &{\mathbf F}^2, \dots, {\mathbf F}^{|{\mathcal F}|}\rangle,\\
\text{where } {\mathbf F}^i &\subseteq S_\ell, 1\leq i \leq |{\mathcal F}|.
\end{split}
\end{equation*}

\item The union of all masks in $\mathcal F$ is the index set.
\end{enumerate}
\begin{equation*}
\bigcup\nolimits_{{\mathbf F}^i \in {\mathcal F}} {\mathbf F}^i = S_\ell.
\end{equation*}
\end{definition}

Take $\ell = 3$ as an example.
The index set $S_3 = \{1,2,3\}$.
Sets such as $\{1\}$, $\{2,3\}$, and $\{1,2,3\}$ are all masks of $S_3$.
$\langle\{1,2\},\{2,3\}\rangle$, $\langle\{2,3\},\{1,2\}\rangle$, and $\langle\{1,2\},\{2,3\},\{1,2\}\rangle$ are three different FOSs of $S_3$.
$\langle \{1,2\},\{2\}\rangle $ is not a valid FOS of $S_3$, since no mask contains $3$.
Note that the masks in FOS are in order, and there may be duplicated masks.



During OM, variables are mixed
according to the elements in FOS.
The operation can be viewed as a variable-wise mask operation,
so elements in FOS are called masks.
For a chromosome $C$, the variables selected according to a mask $M$ are denoted by $C_M$.

Some special FOSs are further defined as follows.

\begin{definition}
For an index set $S_\ell$ and an integer $k$ dividing $\ell$,
a {\bf{homogeneous FOS}} ${\mathcal F}_k$ is defined as
\begin{equation*}
{\mathcal F}_k = \langle{\mathbf F}^1, {\mathbf F}^2, \dots, {\mathbf F}^{\ell / k}\rangle,
\end{equation*}
\begin{equation*}
\text{where } {\mathbf F}^i = \{\pi_j | \left(i-1\right)\cdot k < j \leq i\cdot k \}.
\end{equation*}
\end{definition}

$\vec{\pi} = \langle\pi_1, \pi_2, \dots, \pi_\ell\rangle$ can be any permutation of the index set $S_\ell$.
In other words, ${\mathcal F}_k$ is an FOS consisting of $\ell/k$ disjoint masks, and every mask contains $k$ indexes.
Since the locations of variables should not affect the behavior of optimization,
we use the permutation notation to eliminate the dependency of variable locations.

\begin{definition}\label{def:catFOS}
The {\bf{concatenated homogeneous FOS}} can be defined below.
\begin{equation*}
{\mathcal F}_{k_1, \dots, k_a} = {\mathcal F}_{k_1} \parallel {\mathcal F}_{k_2} \parallel \dots \parallel {\mathcal F}_{k_a},
\end{equation*}
where $\parallel$ is the concatenation operator, which concatenates the elements in the sets while preserving the order.
\end{definition}

Take the case $\ell = 6$ as an example.
$\langle \{1, 5\}, \{2, 4\}, \{3, 6\}\rangle $, 
$\langle \{2, 4, 5\}, \{1, 3, 6\}\rangle $, and
$\langle \{1, 2, 3, 4, 5, 6\}\rangle $
are three valid homogeneous FOSs.
They can be denoted by ${\mathcal F}_2$, ${\mathcal F}_3$, and ${\mathcal F}_6$ respectively.
A valid concatenation of them is
${\mathcal F}_{3,6,2}=\langle \{2,4,5\},\{1,3,6\},\{1,2,3,4,5,6\},\{1,5\},\{2,4\},\{3,6\}\rangle $.

\subsection{Test Problems}
The analyses and experiments are based on three maximization problems.
By successful optimization we mean that the global optimum is found.
The first problem is onemax:
\begin{equation*}
f_{onemax}\left({\vec{x}}\right)=u\left({\vec x}\right),
\end{equation*}
where $u\left({\vec x}\right)$ is the number of $1$s in $\vec x$.
Onemax is trivial, but optimization performance on onemax
indicates how well an algorithm deals with fully separable problems.

The second problem is one instance of the Royal Road functions~\cite{1992_RR},
the structure of which can be characterized by an FOS $\mathcal F$:
\begin{equation}
\label{eq:RR}
f_{royal}\left(\vec{x}\right) = \sum_{\forall {\mathbf F}^i \in {\mathcal F}} R\left(\vec{x}_{{\mathbf F}^i}\right),
\end{equation}
where
\begin{equation*}
R\left(\vec{x}\right) = 
\left\{
\begin{array}{l l}
1 & \text{if }u\left({\vec x}\right)=|{\vec x}|, \\ 
0 & \text{otherwise}.
\end{array}
\right.
\end{equation*}
Note that $\vec{x}_{\mathbf F}$ is the part of $\vec x$ indicated by a mask $\mathbf F$,
and $|\vec x|$ denotes the numbers of elements in $\vec x$.
Motivation of using the Royal Road function is introduced in Section~\ref{sec:nfe}.

The third problem is the deceptive-trap problem~\cite{1992_trap}, which can also be characterized by an FOS $\mathcal F$:
\begin{equation*}
f_{trap}\left(\vec{x}\right) = \sum_{\forall {\mathbf F}^i \in {\mathcal F}} T\left(\vec{x}_{{\mathbf F}^i}\right),
\end{equation*}
where
\begin{equation*}
T(\vec{x}) = 
\left\{
\begin{array}{l l}
1 & \text{if $u(\vec{x}) = |\vec{x}|$}, \\
0.9 \cdot \frac{|\vec{x}|-1-u(\vec{x})}{|\vec{x}|-1} & \text{otherwise}.
\end{array}
\right.
\end{equation*}
The deceptive-trap function is also commonly used for benchmark and is known for its deceptive nature.
Local hill climbing does not lead to the global optimum when solving trap problems.

Note that although these problems are all binary-encoded,
our results can be applied to $\chi$-ary problems.

\subsection{GOMEA}
GOMEA~\cite{2011_OMEA} is one of the first proposed OMEAs and can be integrated with arbitrary FOS.
In this paper, we focus on the variation operator of GOMEA, which is named gene-pool optimal mixing (GOM).
The pseudocode of GOM can be expressed in Algorithm~\ref{alg:GOM}.
Utilizing this operator,
the pseudocode of GOMEA can be expressed in Algorithm~\ref{alg:OMEA}.
The population is denoted by $\mathcal P=\{P_1, \dots, P_n\}$, and
the offspring are denoted by $\{O_1, \dots, O_n\}$.
Optimization halts if the population converges to one instance.
In this paper, all experiments are done with an implementation of GOMEA.
For the experiments that need to determine the minimal required population size, a bisection procedure~\cite{2001_bisection} is adopted.

\begin{algorithm}[h]
\SetKwInOut{Input}{Input}
\SetKwInOut{Output}{Output}
\DontPrintSemicolon
\caption{Genepool Optimal Mixing}
\label{alg:GOM}
\Input{$f$: fitness function, $R$: receiver,\\
${\mathcal P}$: population, ${\mathcal F}$: FOS}
\Output{$O$: offspring}
\BlankLine
$O \gets R$ //{\it Copy to the offspring} \;
$B \gets R$ //{\it Copy to a buffer} \;
\For{$i=1$ to $|{\mathcal F}|$}{
    $r \gets$ random number from $1$ to $|{\mathcal P}|$\;
    $D \gets P_r$ //{\it Select the donor} \;
    $B_{{\mathbf F}^i} \gets D_{{\mathbf F}^i}$ //{\it Get the fragment of the donor} \;
    \If {$f\left(B\right) \geq f\left(O\right)$}{
	$O_{{\mathbf F}^i} \gets B_{{\mathbf F}^i}$ //{\it Adopt the change if improved}\; }
	\Else{
	$B_{{\mathbf F}^i} \gets O_{{\mathbf F}^i}$ //{\it Reset the buffer if not improved} \;
	}
}
\Return $O$
\end{algorithm}

\begin{algorithm}[h]
\SetKwInOut{Input}{Input}
\SetKwInOut{Output}{Output}
\DontPrintSemicolon
\caption{GOMEA}
\label{alg:OMEA}
\Input{$f$: fitness function, $n$: population size,\\
${\mathcal F}$: FOS}
\Output{optimization solution}
\BlankLine
randomly generate $n$ instances for $\mathcal P$\;
\While{$\neg$\textsc{ShouldTerminate}$\left({\mathcal P}\right)$}{
    \For{$i=1$ to $n$}{
	$O_i \gets $\Call{GenepoolOptimalMixing}{$f, P_i, {\mathcal P}, {\mathcal F}$}\;}
    \For{$i=1$ to $n$}{
	$P_i \gets O_i$\;}
}
\Return the best instance in $\mathcal P$
\end{algorithm}

In practice, linkage detection techniques are applied to generate FOS.
In our scenario, this step is omitted.
We manually determine the FOS before optimization to prevent unnecessary noise.
We also disable the function of mutation in all of our experiments,
which is the same as the scenario in the canonical GOMEA.

\subsection{Population Sizing}

For genetic algorithms (GAs), the supply model was proposed based on the concept of 
ensuring an adequate supply of partial solutions.
GAs find the global optimum by mixing the segments of it,
which must exist in the initial population.
Consider a $\chi$-ary problem which can be decomposed into $m$ parts,
and each of them contains $k$ variables.
Since GAs are stochastic processes, we can never guarantee the existence of all partial solutions.
Instead, we consider a tolerable probability $\alpha$ of not having a partial solution in the initial population.
By approximating $\alpha$ by $1/m$, the reciprocal of number of subproblems, 
Goldberg~{\it et~al}.\ derived a minimal required population size in this scenario~\cite{2001_Supply}:
\begin{equation}\label{eq:goldbergSupply}
n=\chi^k\left(k\ln\chi+\ln m\right).
\end{equation}

There are other population-sizing models for GAs.
Goldberg~{\it et~al}.\ proposed one by considering that
larger population leads to higher probability of making correct decisions~\cite{1992_Goldberg}.
Harik~{\it et~al}.\ proposed another model by making an analogy
between selection in GAs and the gambler's ruin problem~\cite{1999_Gamblers}.
These models give different estimations of the required population size.
In this paper, we find out that the supply model is a proper estimator for OMEAs.


%
%


\section{One-Layer Masks}
\label{sec:nonOverlappingMasks}

To begin our study, we first focus on the scenario so-called one-layer masks where FOS consists of disjoint masks.
Three assumptions are made in this section.
First, we only focus on problems which can be fully separated into subproblems.
Second, we assume that the problem structures are perfectly identified before optimization.
Discussion on not well-identified problems is in next section.
Third, we assume the global optimum is unique in every problem.
For problems with multiple global optimums, we expect smaller required population due to smaller need of initial supply.



\subsection{Population Sizing}

The required population size can be estimated from the viewpoint of initial supply.
Consider a set containing $k$ variables which are $\chi$-ary,
in correspondence to a mask with $k$ indexes.
We assume that the optimal state of the set is unique, which is part of the unique global optimum.
A set is considered as correct if it is of optimal state, and is incorrect otherwise.
The probability of the set being correct after
uniformly random initialization is $\chi^{-k}$.
Since the variables are exchanged using the corresponding mask,
they are always tied together during mixing.
Therefore no new pattern is created after mixing.

Since the masks reflect the decomposed subproblems,
correct sets are never replaced,
and eventually dominate the population.
Hence the population converges to the optimal state for the mask
if and only if at least one correct set exists initially.
For a mask of size $k$,
the probability that at least one set of initial variables is correct among the population is
\begin{equation*}
p_{correct} = 1-\left(1-\chi^{-k}\right)^n.
\end{equation*}
EA finds the global optimum if and only if every subproblem contains correct sets initially,
the probability of which is
\begin{equation}\label{eq:pop_size_raw}
p_{success}=\prod\limits_{\forall {\mathbf{F}}^i \in {\mathcal{F}}} \left(1-\left(1-\chi^{-|{\mathbf{F}}^i|}\right)^n\right).
\end{equation}
If the tolerable failure rate of EA is $\alpha$,
solving $p_{success}=1-\alpha$ yields the required population size $n$.

\begin{figure}
\centering
\epsfig{file=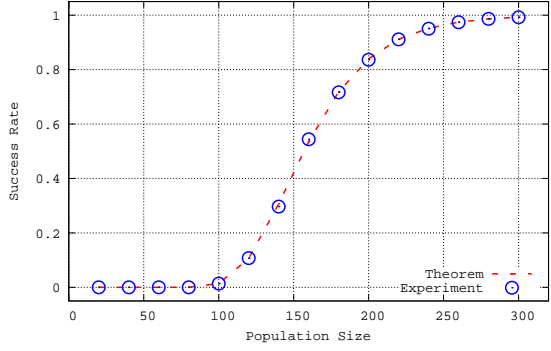, width=3in}
\caption{
Success rate for various population sizes with $\ell =500$ and $k=5$.
Experiments are repeated $10^4$ times.
The maximum absolute error is $1.03\%$.
}
\label{fig:supply}
\end{figure}

A special case of Equation~\ref{eq:pop_size_raw} is that all masks are equal-sized,
which means $|{\mathbf{F}}^i| = k$, $i={1,2, \dots, m}$.
Therefore the equation can be rewritten as
\begin{equation}\label{eq:suc_rate}
1-\alpha=p=\left(1-\left(1-\chi^{-k}\right)^n\right)^m.
\end{equation}
This result is verified empirically in Figure~\ref{fig:supply}.
We can see that the experiment values are accurately estimated.

Equation~\ref{eq:suc_rate} also leads to the required population size:
\begin{equation} \label{eq:pop_size}
n=\frac{\ln\left({1-\left(1-\alpha\right)^{\frac{1}{m}}}\right)}{\ln\left({1-\chi^{-k}}\right)}.
\end{equation}
Note that $0<\frac{1}{m}\leq1$ in Equation~\ref{eq:pop_size}.
To derive the complexity of $n$,
we consider a convex function 
$f\left(x\right)=\left(1-\alpha\right)^x$,
where $x\in(0,1]$.
Two inequalities can be derived:
%
\begin{equation*}
\left(1-x\right)f\left(0\right)+x f\left(1\right)\geq f\left(x\right)\geq f\left(0\right)+x f'\left(0\right)
\end{equation*}
Let $x=\frac{1}{m}$ and take natural log, we have
\begin{equation*}
\frac{\ln\frac{1}{1-\alpha}}{m}\leq 1-\left(1-\alpha\right)^\frac{1}{m} \leq \frac{\alpha}{m},
\end{equation*}
\begin{equation*}
\ln\left( c_1 \frac{1}{m}\right) \geq\ln\left(1-\left(1-\alpha\right)^\frac{1}{m}\right) \geq \ln\left(c_2 \frac{1}{m}\right),
\end{equation*}
where $c_1$ and $c_2$ are some constants.


To compare the optimization cost of problems with different sizes, we assume that the problems have similar structures,
which implies they are encoded in the same way ($\chi$ is constant),
and the adopted masks are similar ($k$ is constant).
The tolerable failure rate $\alpha$ should not vary with the problem size.
By fixing $\alpha$, $\chi$, and $k$,
asymptotically tight bound for Equation~\ref{eq:pop_size} with regarding to $m$ can be derived:

\begin{equation*}
\Theta\left(n\right)=\Theta\left(\ln\left(1-\left(1-\alpha\right)^\frac{1}{m}\right)\right) = \Theta\left(\ln m\right).
\end{equation*}

The complexity matches the supply model in Equation~\ref{eq:goldbergSupply}, which also leads to $n=\Theta\left(\ln m\right)$.
In contrast,
models concerning decision making suggest population sizing from $\Theta\left(\sqrt{m}\right)$ to $\Theta\left(m\right)$~\cite{1992_Goldberg, 1999_Gamblers}.
By comparing the complexities, we verified that the concept of supply suits well for OMEAs, while others do not.

%

\subsection{Convergence Time}

The study of convergence time, the number of generations for OMEAs to converge, involves two parts.
The case of a single mask is first studied,
and the result of which is then extended to two-layer masks.

For the one-mask case,
denote the mask size by $k$ and the generation number by $t$, starting with $t=0$.
$t$ increases 1 after each generation.
Define $p_t$ as the proportion of correct sets among all chromosomes.
$p_0$ can be approximated by its expected value, $\chi^{-k}$.
We estimate the convergence time by modeling the growth of $p_t$.

In GOMEA, two candidates are selected as the donor and receiver,
and the former tries to donate part of its bit pattern to the latter.
The receiver takes the donation only if its fitness does not decrease,
so a correct set never accepts an incorrect one.
On the contrary, incorrect receiver becomes correct once the donor is correct.
Since the expected number of incorrect chromosomes in generation $t$ is $n\left(1-p_t\right)$,
the expected gain in the number of correct sets after one generation is
$n\left(1-p_t\right)\cdot Pr\left(\text{donor is correct}\right)
=n p_t\left(1-p_t\right)$.
Note that changes are made on the offspring instead of receiver,
so $p_t$ remains constant until the end of the current generation.
Thus we get the iterative equation:
\begin{equation}
\label{eq:p_ite}
n p_{t+1}-n p_t = n\cdot p_t\cdot\left(1-p_t\right).
\end{equation}
For $t\in{\mathbb N}_0 = {\mathbb N} \cup \{0\}$,
denote $1-p_t$ by $q_t$, and we have
\begin{equation*}
q_t-q_{t+1} = \left(1-q_t\right)q_t,
\end{equation*}
\begin{equation*}
q_t^2 =  q_{t+1}.
\end{equation*}
Along with $q_0=1-p_0=1-\chi^{-k}$, we get
\begin{equation} \label{eq:p_t}
p_t = 1-q_t=1-\left(1-\chi^{-k}\right)^{2^t}, t\in{\mathbb N}_0.
\end{equation}

Although the solution to Equation~\ref{eq:p_ite} is derived,
no $t$ satisfies $p_t=1$,
because we approximate $p_t$ by its expected value.
Since $p_t$ represents the proportion of correct instances in the population with $n$ candidates,
the greatest value of $p_t$ less than $1$ is $1-1/n$. 
Since the expected time for $p_t$ to reach $1-1/n$ is less than the convergence time,
we derive a lower bound for the latter by solving $p_{t_{conv}} > 1-1/n$.

To estimate the lower bound $t_L$,
we approximate $p_t$ by a continuous function
$p\left(t\right)$, which treats $t$ as a real number.
From Equation~\ref{eq:p_t}, we set
\begin{equation}
p\left(t\right) = 1 - q\left(t\right) = 1 - \left(1-\chi^{-k}\right)^{2^t}, t \geq 0.
\end{equation}
Note that $p\left(t\right) = p_t$ for $t\in{\mathbb N}_0$.
Solving $p\left(t_{conv}\right)>1-\frac{1}{n}$ yields
\begin{equation} \label{eq:conv1}
t_{conv} > t_L = \log_2 \left(\frac{\ln\left( \frac{1}{n}\right)}{\ln\left(1-\chi^{-k}\right)}\right).
\end{equation}


%


The result can be further extended to multiple disjoint masks.
Since the problem is separable,
convergence in one position does not interfere with that in another position.
Convergence time is thus dominated by the variable set with the longest convergence time.

Assume that the chromosome consists of $m$ sets of variables,
convergence time of which are
$t_{conv,1}, \dots, t_{conv,m}$ respectively.
We have $t_{conv} = \max_i\left({t_{conv,i}}\right)$.
According to Equation~\ref{eq:conv1},
smaller initial proportion leads to longer convergence time.
For the case that each set contains $k$ variables,
we can approximate the maximum of $t_{conv,i}$ by finding the convergence time of the set with
least initial correct instances.
Let $X_1, \dots, X_m$ be $m$ independent and identically distributed random variables following the binomial distribution $B\left(n,\chi^{-k}\right)$.\footnote
{
$B\left(n,p\right)$ denotes the distribution with the
pmf $Pr\left(x\right)=
\left\{
\begin{array}{l l}
\frac{n!}{\left(n-x\right)!x!} p^x \left(1-p\right)^{n-x} & x\in\{0,1,\dots,n\}, \\
0 & \text{otherwise}.
\end{array}
\right.$
}
Denote the first order statistic by $X_{\left(1\right)}=\min_i\left(X_i\right)$. Let $E[X_{\left(1\right)}] = x_{\left(1\right)}$.
Hence for multiple masks we have
\begin{equation}
\label{eq:conv}
t_{conv} > t_L = 
\log_2 \left(\frac{\ln \left(\frac{1}{n}\right)}{\ln\left(1-\frac{x_{\left(1\right)}}{n}\right)}\right).
\end{equation}

\begin{figure}
\centering
\epsfig{file=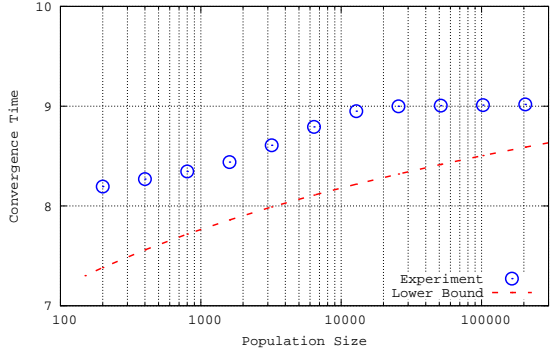, width=3in}
\caption{
Convergence time with one mask of size $5$.
For population size ranging from $200$ to $204800$,
the maximum difference between experiment values and the lower bound is $0.81$ generations.
The difference decreases for large population.
}
\label{fig:conv1}
\end{figure}

\begin{figure}
\centering
\epsfig{file=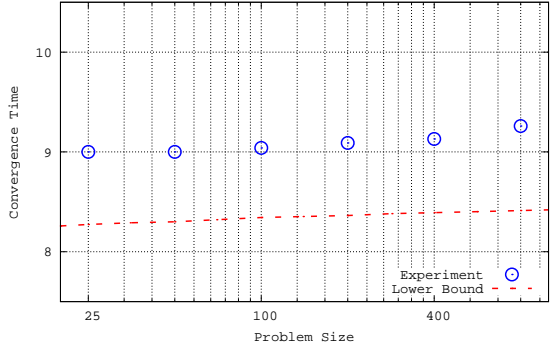, width=3in}
\caption{
Convergence time for $n=10000$.
For problem size ranging from $25$ to $800$,
the maximum difference is $0.85$ generations.
}
\label{fig:conv2}
\end{figure}

\begin{figure}
\centering
\epsfig{file=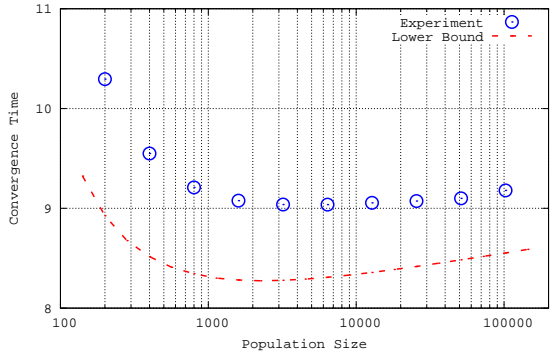, width=3in}
\caption{
Convergence time for $\ell=100$ with masks of size 5.
For population size ranging from $200$ to $102400$,
the maximum difference between experiment values and the lower bound is $1.37$ generations.
The difference decreases for large population.
}
\label{fig:conv3}
\end{figure}

The lower bound we derived is verified with results on the onemax problem.
Figure~\ref{fig:conv1} shows the case of one mask, where the lower bound is from Equation~\ref{eq:conv1}.
The case of multiple disjoint masks is shown in Figures~\ref{fig:conv2} and \ref{fig:conv3},
where the lower bound is from Equation~\ref{eq:conv}.

The proposed lower bound strictly bounds the empirical results.
For large $n$, $1-1/n$ approaches $1$.
Since the lower bound is derived by solving $p\left(t_L\right) = 1-1/n$,
the lower bound should be close to the convergence time for large population.
This can be verified in Figures~\ref{fig:conv1} and \ref{fig:conv3}.
Since an almost converged population is likely to converge in one more generation,
the lower bound is about one generation below experiment results.

In the case of multiple masks,
we find out that the convergence time decreases while $n$ increases when $n$ is small.
This is mainly because of the $\frac{x_{\left(1\right)}}{n}$ term in Equation~\ref{eq:conv}.
According to the central limit theorem,
the first order statistics of the proportion of correct instances increases and approaches $\chi^{-k}$ when $n$ grows.
When $n$ is large, the $\ln{\left(\frac{1}{n}\right)}$ term dominates the lower bound, making it increase when $n$ grows.
This phenomenon is verified in Figure~\ref{fig:conv3}.

\subsection{Number of Function Evaluations}\label{sec:nfe}

The function evaluations of OMEAs consist of two parts.
The first is those during the initialization, and the second is those during OM.
Therefore the total NFE is
\begin{equation}
n_{fe} = n + n_{fe,OM}.
\end{equation}
We then model $n_{fe,OM}$ for the case of a single size-$k$ mask.

As before, the proportion of correct subsolutions in the population at generation $t$ is $p_t$.
Under the assumption of unique global optimum,
$\chi^k-1$ subsolutions are not optimal.
Denote each of their densities in the population by $q^{(1)}_t, \dots, q^{(\chi^k-1)}_t$, respectively.
In this generation, a receiver only needs to be evaluated when it differs from the donor.
Therefore the probability that a receiver needs the evaluation is
\begin{equation}\label{eq:e_t}
e_t = 1-p_t^2-\left(q^{\left(1\right)}_t\right)^2-\left(q^{\left(2\right)}_t\right)^2-\dots-\left(q^{\left(\chi^k-1\right)}_t\right)^2.
\end{equation}
We already have the approximation of $p_t$ in Equation~\ref{eq:p_t}.
Since $q^{(1)}_t, \dots, q^{(\chi^k-1)}_t\in [0,1]$, we have
\begin{equation*}
\frac{\left(\sum\limits_{i=1}^{\chi^k-1} q_t^{\left(i\right)} \right)^2}{\chi^k-1}
    \leq \sum\limits_{i=1}^{\chi^k-1} \left(q_t^{(i)}\right)^2
    \leq \left(\sum\limits_{i=1}^{\chi^k-1} q_t^{(i)}\right)^2.
\end{equation*}
With the condition that
\begin{equation*}
\sum\limits_{i=1}^{\chi^k-1} q_t^{(i)} = 1-p_t,
\end{equation*}
Equation~\ref{eq:e_t} leads to
\begin{equation}\label{eq:UL}
1-p_t^2-\frac{\left(1-p_t\right)^2}{\chi^k-1}
\geq e_t
\geq 1-p_t^2-\left(1-p_t\right)^2.
\end{equation}

The required NFE can be approximated by
\begin{equation*}
n_{fe,OM} = \sum\limits_{t=0}^{\infty} n\cdot e_t  = n \sum\limits_{t=0}^{\infty} e_t.
\end{equation*}
By expanding the summation, we can derive the upper and lower bounds of $n_{fe}$.
For the upper bound, there is no closed form.
Since 
$\sum\limits_{t=0}^{\infty}\left(1-p_t^2-\frac{\left(1-p_t\right)^2}{\chi^k-1}\right)$ is a function of $k$ if $p_t$ is approximated by Equation~\ref{eq:p_t},
we denote it by a function $U\left(k\right)$,
where $U\left(k\right)$ is a function from $\mathbb{N}$ to $\mathbb{R}$.
$\mathbb{N}$ is the natural number set, and $\mathbb{R}$ is the real number set.
Here the variable $\chi$ is considered constant.
So we have
\begin{equation*}
n_{fe,OM} \leq n\cdot U\left(k\right).
\end{equation*}

There is a closed form of the lower bound:
\begin{equation*}
\begin{aligned}
\sum\limits_{t=0}^{\infty}\left(1-p_t^2-\left(1-p_t\right)^2\right)
=\sum\limits_{t=0}^{\infty} \left(2 q_t - 2 q_t^2\right) \\
=\sum\limits_{t=0}^{\infty} \left(2q_t - 2q_{t+1}\right)
=2q_0.
\end{aligned}
\end{equation*}
This yields
\begin{equation*}
n_{fe,OM} \geq n\cdot 2q_0 = n\cdot2\left(1-\chi^{-k}\right) = n\cdot L\left(k\right),
\end{equation*}
where $L\left(k\right) = 2\left(1-\chi^{-k}\right)$.
Hence the NFE can be bounded by
\begin{equation*}
n\cdot U\left(k\right) \geq n_{fe,OM} \geq n\cdot L\left(k\right).
\end{equation*}

The results can be extended to arbitrary number of masks, $m$.
Since all of our masks are disjoint, evaluations of them are independent to each others.
So the total NFE is the summation of all required NFEs for each mask.
Therefore we get
\begin{equation*}
n\cdot \sum_{\forall {\mathbf{F}}^i \in {\mathcal{F}}} U\left(|{\mathbf{F}}^i|\right)
\geq
n_{fe,OM}
\geq
n\cdot \sum_{\forall {\mathbf{F}}^i \in {\mathcal{F}}} L\left(|{\mathbf{F}}^i|\right),
\end{equation*}
or equivalently,
\begin{equation}
n+n\cdot \sum_{\forall {\mathbf{F}}^i \in {\mathcal{F}}} U\left(|{\mathbf{F}}^i|\right) 
\geq
n_{fe}
\geq
n+n\cdot \sum_{\forall {\mathbf{F}}^i \in {\mathcal{F}}} L\left(|{\mathbf{F}}^i|\right).
\end{equation}
For the special case that the sizes of all masks are equal, we have
\begin{equation}
\label{eq:nfe}
n\cdot\left(1+mU\left(k\right)\right)
\geq
n_{fe}
\geq
n\cdot\left(1+mL\left(k\right)\right),
\end{equation}
where $k$ is the size of every mask,
and $m$ is the number of masks.

The problem nature affects the required number of evaluations.
In Equation~\ref{eq:nfe}, the equality of upper bound holds when $q_t^{\left(1\right)} = q_t^{\left(2\right)} = \dots = q_t^{\left(\chi^k-1\right)}$.
Consider one of our test problems, the Royal Road function. 
By definition, all suboptimals in one subproblem (the $R\left(\vec x\right)$ in Equation~\ref{eq:RR}) contribute equally to the fitness.
Because all suboptimals are equally competitive during OM,
proportions of all suboptimal instances are roughly equal.
Since the equality of upper bound holds,
this problem costs more function evaluations than others do.

\begin{figure}
\centering
\epsfig{file=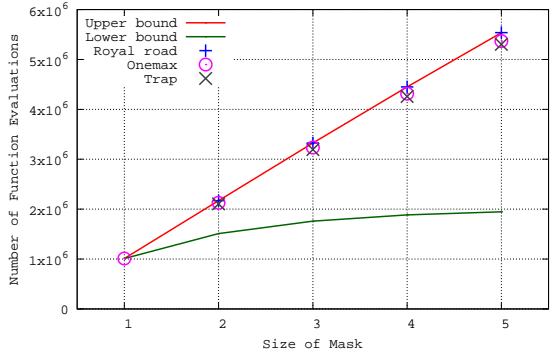, width=3in}
\caption{
$n_{fe}$ with different sizes of masks.
Each FOS contains $100$ masks, and the population size is fixed at $10^4$.
Note that the Royal Road, the onemax, and the trap function are identical for $k=1$.
}
\label{fig:nfe_k}
\end{figure}

\begin{figure}
\centering
\epsfig{file=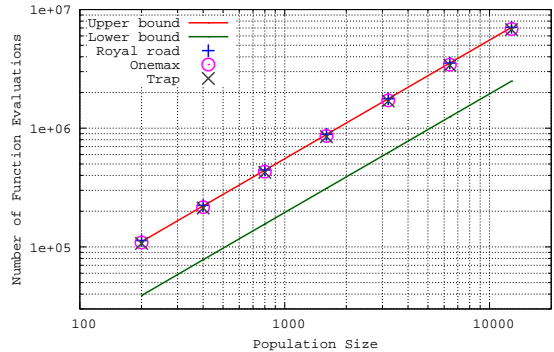, width=3in}
\caption{$n_{fe}$ with different sizes of population.
$m=100$ and $k=5$ for the test problems.
}
\label{fig:nfe_n}
\end{figure}

The results are verified in the following experiments.
Figure~\ref{fig:nfe_k} shows NFE with masks of various sizes.
Figure~\ref{fig:nfe_n} shows the result with various population sizes.
The theoretical values are from Equation~\ref{eq:nfe}.

\begin{figure}
\centering
\epsfig{file=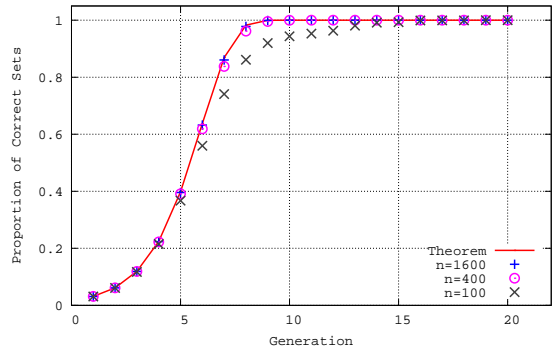, width=3in}
\caption{$p_t$ for various population sizes.
The problem is a $500$-bit onemax problem, with masks of size $5$.
We can see smaller $n$ leads to slower convergence.}
\label{fig:pt_n}
\end{figure}

In these two experiments, we have made some observations.
First, the lower bound is generally far from experiment results when there are multiple suboptimals,
since we derive the lower bound by ignoring the cost of comparison between suboptimals
in Equation~\ref{eq:UL}.
Second,
NFE of the Royal Road function exceeds the upper bound by $2.17\%$ when $n=200$ and by $0.02\%$ when $n=12800$
in Figure~\ref{fig:nfe_n}.
We believe that the reason is as follows.
The estimation of $p_t$ deriving from the expected value 
while $p_t$ is discrete is the reason.
For small $n$, the approximation is inaccurate.
This is verified in Figure~\ref{fig:pt_n}, which indicates that
smaller population leads to slower convergence, 
implying additional consumption in function evaluations.
This explains why NFE is slightly underestimated, especially when $n$ is small.
Third, NFE of the Royal Road problem is always the greatest,
and NFE of the trap problem is always the least.
This verifies that the subproblem affects the NFE, yet the difference is subtle.
Last, the upper bound seems a proper estimator for the NFE for all test problems, 
with relative error less than $5\%$.
We infer that the high mask-wise selection pressure of OM
quickly filters out the incorrect sets,
and therefore the convergence behavior is not much affected from the subproblem structures.

\section{Two-layer Masks}
\label{sec:twoLayerNonOverlappingMasks}

In this section, we extend our study to two-layer masks as the first step to multi-layer masks.
Multi-layer masks are usually adopted since in practice most problem structures are unknown and not fully separable.
In this section, we focus on the onemax problem with FOS in the form ${\mathcal F}_{k,1}$ with $k>1$. 
Note that the FOS is defined in Definition~\ref{def:catFOS}.


The population converges to the global optimum if and only if all the variables converge to $1$s.
For an index $i$, consider all the variables $\vec{x}_i$ in every chromosomes. 
There are $n$ variables with this index.
We use the notation $p$ to denote the proportion of $1$s in these $n$ variables.

Since masks in a homogeneous FOS do not share common variables,
exactly one mask in ${\mathcal F}_k$ and one mask in ${\mathcal F}_1$ contain $i$.
When mixing with one-bit masks, $1$s never change to $0$s,
but this is not the case of $k$-bit masks.
Since the fitness is evaluated when all $k$ variables are exchanged,
the overall fitness never decreases, but some variables can be ruined.
For example, the pattern {\tt 101} overwrites {\tt 010} when a 3-bit mask applies to them, 
since the fitness increases by 1.
However, the second variable in {\tt 010}, which is one, is overwritten by a zero.
For a certain index, if all $n$ variables with the index in the population become $0$s,
the optimization fails.

This phenomenon can be further quantified from the viewpoint of probability.
By the assumption of random initialization, $p=2^{-1}$ in binary-coded problems.
For a set of $k$ variables covered by a $k$-bit mask,
the number of $1$s in the set can be expressed as a random variable.
If the first variable is zero, the random variable $X_0$ follows the Binomial distribution $B(k-1,2^{-1})$.
Recall that $k$ is an integer greater than one.
If the first variable is one, the random variable is $X_1=1+X_0$.
Consider a variable ${\vec x}_i$ equals to $0$ in the donor and equals to $1$ in the receiver.
${\vec x}_i$ in the receiver becomes $0$ with probability $Pr\left(X_1\leq X_0\right)$,
which we called the reverse-growth probability, $p_{rg}$.
The probabilities are calculated using the Binomial distribution and shown in Table~\ref{tbl:x1x0}.
We can see larger $k$ leads to larger $p_{rg}$. 

\begin{table}
\centering
\begin{tabular}{|c|c|c|c|c|} \hline
$k$ & $Pr(X_1>X_0)$ & $Pr(X_1=X_0)$ & $Pr(X_1<X_0)$ & $p_{rg}$\\ \hline
2 & 75\% & 25\% & 0\%   & 25\%\\
3 & 69\% & 25\% & 6\%   & 31\%\\
4 & 66\% & 23\% & 11\%  & 34\%\\
5 & 64\% & 22\% & 14\%  & 36\%\\ \hline
\end{tabular}
\caption{Comparison of two random variables $X_1$ and $X_0$.
Larger $k$ yields larger $p_{rg}$, which is the probability that $X_1 \leq X_0$.
}
\label{tbl:x1x0}
\end{table}

The required population size varies when adopting various FOSs.
The concept of cross competition explains this phenomenon~\cite{1992_CC}.
Cross competition is first introduced to derive an upper bound for the selection pressure in GA.
To prevent cross-competitive failure, the selection pressure must satisfy
\begin{equation*}
s<n\frac{\ln\left(1-p_0\right)}{\ln \alpha},
\end{equation*}
where $s$ is the selection pressure, $p_0$ is the probability that a correct bit is preserved into the next generation,
and $\alpha$ is a threshold, which means optimization is likely to fail when $\alpha\ell$ variables are incorrectly converged.
After some modification, we obtain
\begin{equation}\label{eq:CC}
n>\frac{s\cdot \ln \alpha}{\ln\left(1-p_0\right)}.
\end{equation}

Although in the scenario of GOMEA, the situation is different, but the similar concept applies.
Larger $k$ leads to larger $p_{rg}$,
which then leads to lower probability that a correct bit is preserved.
In other words, larger $k$ yields smaller $p_0$.
By assuming that $s$ and $\alpha$ do not vary much with $k$,
the RHS of Equation~\ref{eq:CC} increases when $k$ increases.
This means that the lower bound of population size increases.
In other words, the required population increases if $k$ increases when adopting ${\mathcal F}_{k,1}$.

\begin{figure}
\centering
\epsfig{file=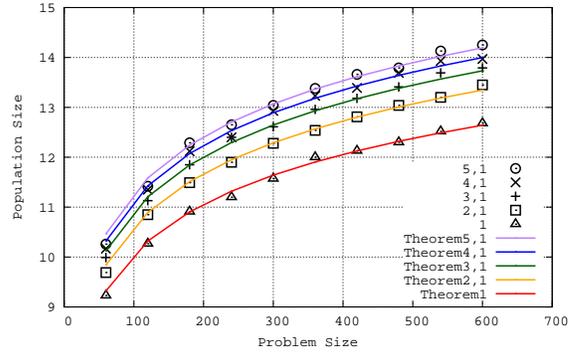, width=3in}
\caption{Required population size for various FOS.
Theoretical values of ${\mathcal F}_{k,1}$
are estimated by curve fitting as multiple of that of ${\mathcal F}_1$.
The maximum relative error among all estimations is $2.0\%$.
}
\label{fig:pop_k1}
\end{figure}

The following experiments are conducted to verify the above hypothesis.
For various lengths of problems, different FOSs are adopted to solve the onemax problem.
These FOSs are ${\mathcal F}_{5,1}$, ${\mathcal F}_{4,1}$, ${\mathcal F}_{3,1}$, ${\mathcal F}_{2,1}$, and ${\mathcal F}_1$.
The result is shown in Figure~\ref{fig:pop_k1},
indicating that optimization using ${\mathcal F}_{k,1}$ with larger $k$ requires larger population.
Furthermore,
a proportional relationship exists between the required population size for ${\mathcal F}_{k,1}$ and ${\mathcal F}_1$,
while the latter can be approximated by Equation~\ref{eq:pop_size}.
From this observation,
the theoretical values of required population for ${\mathcal F}_{k,1}$ are obtained by fitting the experiment values to a constant multiple of that of ${\mathcal F}_1$.
According to the results, masks which cover multiple separable subproblems are not beneficial from the viewpoint of population size.

\section{Conclusion}
\label{sec:conclusion}

In this paper, we derived the convergence models of OMEAs
for one-layer and two-layers masks.
For problems with separable structures,
behaviors of GOMEA with masks corresponding to the structures were analyzed.
The required population size was accurately estimated from the viewpoint of initial supply.
Analyzing the growth of sub-solution led to the convergence-time model.
The NFE was then estimated.
These three models were verified empirically,
and the values of relative error among all experiments were less than $5\%$.
For multi-layer disjoint masks, a special case with two-layer masks was studied.
We found that the concept of cross competition explains the growth in the required population sizes.

As for future work, we would like to extend our results to multi-layer masks by quantifying the effect of cross competition and by investigating the growth of subsolutions iteratively.
Hopefully, we would be able to analyze the behavior of OMEAs with full linkage tree,
which is a special case of multi-layer FOS.
Furthermore, since this paper is limited by the assumptions we made, we would like to study the case with weaker assumptions.


The major contributions of this paper reside in the derivations of
the population-sizing, the convergence-time,
and the NFE models for OMEAs with one-layer and two-layer masks.
Our models are empirically verified and the relative errors of our estimators are small.
In addition, our models lead to the following insightful findings.
First, for the case of one-layer masks, the required population size is decided by initial supply rather than decision making.
This explains why OMEAs generally require relatively small populations compared to EDAs.
Second, NFE for the test problems is very close to the proposed upper bound,
by which we infer that the mask-wise selection pressure of OM quickly filter out non-optimum subsolutions, making the subproblem composition insignificant.
Third, for two-layer masks,
the required population size is proportional to that of the one-layer masks,
and the ratio is positively related to the reverse-growth probability.


\section{ACKNOWLEDGMENTS}

The authors would like to thank the support by Ministry of Science and Technology in Taiwan under Grant No.\ MOST 103-2221-E-002-177-MY2-1.

\bibliographystyle{abbrv}

\end{document}